\newcommand{\ignore}[1]{}

\documentclass[9pt, conference]{IEEEtran}

\usepackage{pdfpages}
\usepackage{listings}
\usepackage{url}
\usepackage{multirow}
\usepackage{subfigure}
\usepackage{epsfig}
\usepackage{graphicx}
\usepackage{amsmath}
\usepackage{amssymb}
\usepackage{color}
\usepackage{setspace}
\usepackage{cite}
\usepackage{hhline}   
\usepackage{algorithm}
\usepackage{algpseudocode}
\usepackage{flushend}
\usepackage{url}
\usepackage[normalem]{ulem}
\usepackage{graphics}
\usepackage{tikz}
\usepackage{array}
\usepackage{animate}
\usepackage{titlesec}
\usepackage{balance}
\usepackage{enumitem}
\usepackage{booktabs}
\usepackage{hyperref}
\hypersetup{
  hidelinks,
  colorlinks,
  linkcolor={red!80!black},
  citecolor={blue!80!black},
  urlcolor={blue!80!black}
}

\titlespacing*{\section}{0pt}{3pt}{3pt}
\titlespacing*{\subsection}{0pt}{3pt}{2pt}
\titlespacing*{\subsubsection}{0pt}{3pt}{2pt}

\definecolor{codegreen}{rgb}{0,0.6,0}
\definecolor{codegray}{rgb}{0.5,0.5,0.5}
\definecolor{codepurple}{rgb}{0.58,0,0.82}
\definecolor{backcolour}{rgb}{0.95,0.95,0.92}
\definecolor{Burgundy1}{RGB}{159,29,53}
\lstdefinestyle{mystyle}{
    commentstyle=\color{codegreen}\itshape,
    keywordstyle=\color{black},
    numberstyle=\tiny\color{codegray},
    stringstyle=\color{Burgundy1},
    basicstyle=\scriptsize,
    breakatwhitespace=false,         
    breaklines=true,                 
    captionpos=b,                    
    keepspaces=true,                 
    backgroundcolor=\color{backcolour},   
    numbersep=2pt,                  
    showspaces=false,                
    showstringspaces=false,
    showtabs=false,                  
    tabsize=2
}

\begin{document}


\title{DALI-PD: Diffusion-based Synthetic Layout Heatmap Generation for ML in Physical Design}

\author{
    Bing-Yue Wu,
    Vidya A. Chhabria \\
    \IEEEauthorblockA{Arizona State University, Tempe, AZ, USA}
}

\author{Bing-Yue Wu and Vidya A. Chhabria \\  Arizona State University}

\maketitle

\begin{abstract}
Machine learning (ML) has demonstrated significant promise in various physical design (PD) tasks. However, model generalizability remains limited by the availability of high-quality, large-scale training datasets. Creating such datasets is often computationally expensive and constrained by IP. While very few public datasets are available, they are typically static, slow to generate, and require frequent updates. To address these limitations, we present DALI-PD, a scalable framework for generating synthetic layout heatmaps to accelerate ML in PD research. DALI-PD uses a diffusion model to generate diverse layout heatmaps via fast inference in seconds. The heatmaps include power, IR drop, congestion, macro placement, and cell density maps. Using DALI-PD, we created a dataset comprising over 20,000 layout configurations with varying macro counts and placements. These heatmaps closely resemble real layouts and improve ML accuracy on downstream ML tasks such as IR drop or congestion prediction. 
\end{abstract}

\section{Introduction} 
\label{sec:intro}
Over the past decade, machine learning (ML) has been increasingly applied to various physical design (PD) tasks, including floorplanning, placement, routing, IR drop, congestion prediction, and timing optimization~\cite{barboza2019machine,mlinsight,cheng2020fast,guo2022time,2025iccadgenmethodineda,2025aspdacpathgen,chhabria2022from,Kahng_Nerem_Wang_Yang_2024,chhabria2025strengthening,2022mlcadwu,2022daclbr,circuitops}. As ML models grow more sophisticated, their performance hinges on the availability of high-quality, diverse, and large-scale training datasets~\cite{iscas85,itc99,ibm-power-grid,ozdal2012ispd,ozdal2013ispd,2023circuitnet,2024circuitnet,edacorpus,reddy2019machine,2024cadcontestc,ispd2019benchmark}, However, such data remains scarce. Unlike fields like computer vision or natural language processing, where massive labeled datasets are readily available~\cite{imagenet,liu2015faceattributes,2022nipslaion-5b,vanetten2021multitemporalurbandevelopmentspacenet}, PD datasets are typically proprietary, lack diversity, and are expensive to generate—in terms of both manual effort and computational cost. This poses a challenge in developing generalizable ML models.

To support EDA research, the community has introduced several benchmarks such as ISCAS’85~\cite{iscas85}, ITC’99~\cite{itc99} for logic synthesis and test, IBM power grid benchmarks~\cite{ibm-power-grid} for IR drop, ISPD 2012/2013, ICCAD 2024~\cite{ozdal2012ispd, ozdal2013ispd, 2024cadcontestc} for logic gate sizing, and MacroPlacement~\cite{macroplacementpaper,tilos-macroplacement}. While impactful, these benchmarks are limited in scale and diversity, making them insufficient for ML applications that require extensive and varied datasets.

More recently, ML-specific datasets such as CircuitNet~\cite{2023circuitnet,2024circuitnet} offering labeled layout data,  EDA Corpus~\cite{edacorpus} offering data to train large-language models for PD tasks, or VerilogEval~\cite{VerilogEval} offering large dataset for RTL generation.  However, these datasets are typically static—released once and seldom updated—due to the significant manual and computational effort required~\cite{2023circuitnet}. Datasets released by industry are often obfuscated to protect IP, which can distort data distributions and reduce their utility~\cite{ibm-power-grid, ozdal2012ispd, ozdal2013ispd}. The overhead involved in creating and maintaining such datasets makes continuous updates impractical. To address these limitations, researchers have begun exploring automated generation of synthetic datasets. Tools like the artificial netlist generator (ANG) \cite{kim2021machine} aim to generate synthetic netlists that still require running place and route tools for layout data.

Different ML tasks in PD require different data modalities, such as image-like heatmaps for IR drop or congestion, and graphs for timing analysis. Heatmaps are especially common for representing spatial PD characteristics such as power density prediction~\cite{GNN}, IR drop prediction~\cite{maveric, powernet, iredge}, DRC violation prediction~\cite{2018userudy2}, etc. Our focus in this work is on generating synthetic heatmaps for PD.  Advances in image generation—such as generative adversarial networks (GANs) and diffusion models—have achieved impressive results in domains like face synthesis and text-to-image generation~\cite{10463372}. Building on this progress, BeGAN~\cite{2021began} applied GANs to generate synthetic power maps. However, BeGAN is limited to generating current maps only of fixed sizes, and lacks the spatial fidelity required for realistic, high-resolution PD datasets that span multiple applications such as IR drop, congestion, and macro placement.

In this work, we present DALI-PD (inspired by DALL·E image generator from OpenAI~\cite{DALLE}), a scalable framework for synthesizing high-resolution layout heatmaps for multiple physical design applications. DALI-PD is a \underline{d}iffusion-b\underline{a}sed synthetic \underline{l}ayout heatmap generator for ML appl\underline{i}cations in  \underline{PD}.  DALI-PD leverages a two-stage generative pipeline consisting of a variational autoencoder (VAE) for layout representation learning and a UNet-based diffusion model for high-fidelity heatmap generation. Our framework enables fast inference, producing new data points in seconds, while capturing the complex statistical and spatial characteristics of real layouts. DALI-PD generates diverse heatmaps of layouts, including power, IR drop, congestion, macro, and cell density distributions.

\noindent
The key contributions of DALI-PD are: 
\begin{enumerate}
    \item DALI-PD is the first work to apply diffusion models for synthetic data generation for applications of ML in PD. 
    \item Via a fast ML inference, DALI-PD can generate synthetic diverse layout heatmaps in a few seconds.
    \item DALI-PD can generate layouts with different area dimensions, aspect ratio settings, different numbers of macros, utilizations, and clock period. 
    \item Using DALI-PD, we create a large synthetic yet realistic dataset of over 20,000 layout data points. These vary in layout area from 500$\mu$m × 500$\mu$m to 1200$\mu$m × 1200$\mu$m, number of macros from 1 to 50, aspect ratio from 1 to 2, and utilization from 60\% to 90\%, making the diverse dataset suitable for training ML models for a variety of applications in PD.
    \item DALI-PD dataset shares similar characteristic trait as real layouts implemented by EDA tools and can generate these heatmaps in a few seconds compared to the hours taken by EDA tools. 
\end{enumerate}

The large synthetic DALI-PD dataset can be found at~\cite{DALI-PD-GH-commit}.
\section{Related work}
\label{sec:preliminaries}


\noindent
{\bf Existing datasets for ML (CircuitNet)}
ML-oriented datasets such as CircuitNet~\cite{2023circuitnet, 2024circuitnet} have been released to support data-driven research in PD. CircuitNet provides a large number of data points (layout heatmaps) generated from six unique RTL designs by varying EDA tool configurations, clock periods, and utilization targets during placement and routing. This has created over $10,000$ different layout configurations with real circuit heatmaps for the $28$nm technology node. All heatmaps are generated using commercial tools, requiring hours of runtime per data point. Their dimensions range from $222$ to $314$ pixels, where each pixel represents $2.25\mu m$. However, the dataset's diversity is fundamentally limited as acquiring RTLs is challenging, and each data point requires several hours to generate. As a result, both the scale and variety of data remain constrained.  

CircuitNet contains six types of layout heatmaps: (1) cell density, (2) macro regions, (3) RUDY~\cite{2007rudy}, (4) IR drop, (5) power, and (6) toggle rate-scaled power~\cite{powernet}. Cell density maps show cell counts per unit area, while macro maps highlight regions occupied by macros. RUDY maps estimate routing demand post-placement and are widely used for congestion prediction~\cite{2021userudy1,2018userudy2}. IR drop maps capture voltage drops across the chip. Power maps combine leakage, switching, and internal power, while toggle rate-scaled power maps scale dynamic power by toggle rate. These heatmaps support various ML applications, including IR drop, congestion, and DRC prediction.

\noindent
{\bf ML for synthetic data generation (BeGAN)} 
In the broader AI community, synthetic data is widely used to overcome the scarcity of real-world datasets~\cite{nikolenko2021synthetic}. This approach holds similar promise for ML in PD, where access to real circuit data is limited. GANs and diffusion models have successfully generated realistic synthetic data across domains, improving downstream performance~\cite{10822459,10229866,tosato2023eegsyntheticdatageneration,10550673}. For example, diffusion models have enhanced drug discovery~\cite{10822459}, pedestrian detection~\cite{10229866}, EEG-based interfaces~\cite{tosato2023eegsyntheticdatageneration}, and 3D human pose estimation~\cite{10550673}. In the context of PD, BeGAN~\cite{2021began}, inspired by facial image synthesis~\cite{shen2018facefeatgantwostageapproachidentitypreserving}, used GANs and transfer learning to generate synthetic power grid current maps. It leveraged the structural similarity between chip layouts and urban satellite imagery, pretraining on the latter and fine-tuning on limited circuit data to produce realistic synthetic benchmarks.

\noindent
\textbf{DALI-PD.} Our proposed approach, DALI-PD, inspired by BeGAN, also uses AI to generate synthetic layout heatmap distributions and leverages the visual similarity between chip layouts and urban satellite imagery.  However, it introduces several key differences:

\begin{enumerate}
    \item While BeGAN focuses on generating synthetic power maps only, DALI-PD generates all six categories of heatmaps (Fig.~\ref{fig:example}).
    
    \item  BeGAN generates fixed-size maps, whereas DALI-PD’s diffusion model generates layouts of varying sizes, macro configurations, utilization levels, and clock periods (increasing diversity).
    
    \item  BeGAN relies on GANs, while DALI-PD leverages a diffusion model, pretrained on satellite images~\cite{2024diffusionsat},  to generate high-resolution, spatially consistent heatmaps in the six categories.
\end{enumerate}

Unlike CircuitNet, which requires slow EDA runs to generate each heatmap, DALI-PD produces synthetic heatmaps in seconds using fast ML inference while preserving key characteristics of real CircuitNet data. It fine-tunes a diffusion model on CircuitNet’s designs to create a larger, diverse, and realistic dataset. Moreover, CircuitNet is limited to six designs with fixed settings and aspect ratios. DALI-PD expands this dataset efficiently—without costly EDA tool runs.


\section{DALI-PD Framework}
\label{sec:method}
\begin{figure}[t]
    \centering
    \vspace{-6mm}
    \includegraphics[width= 0.87\linewidth]{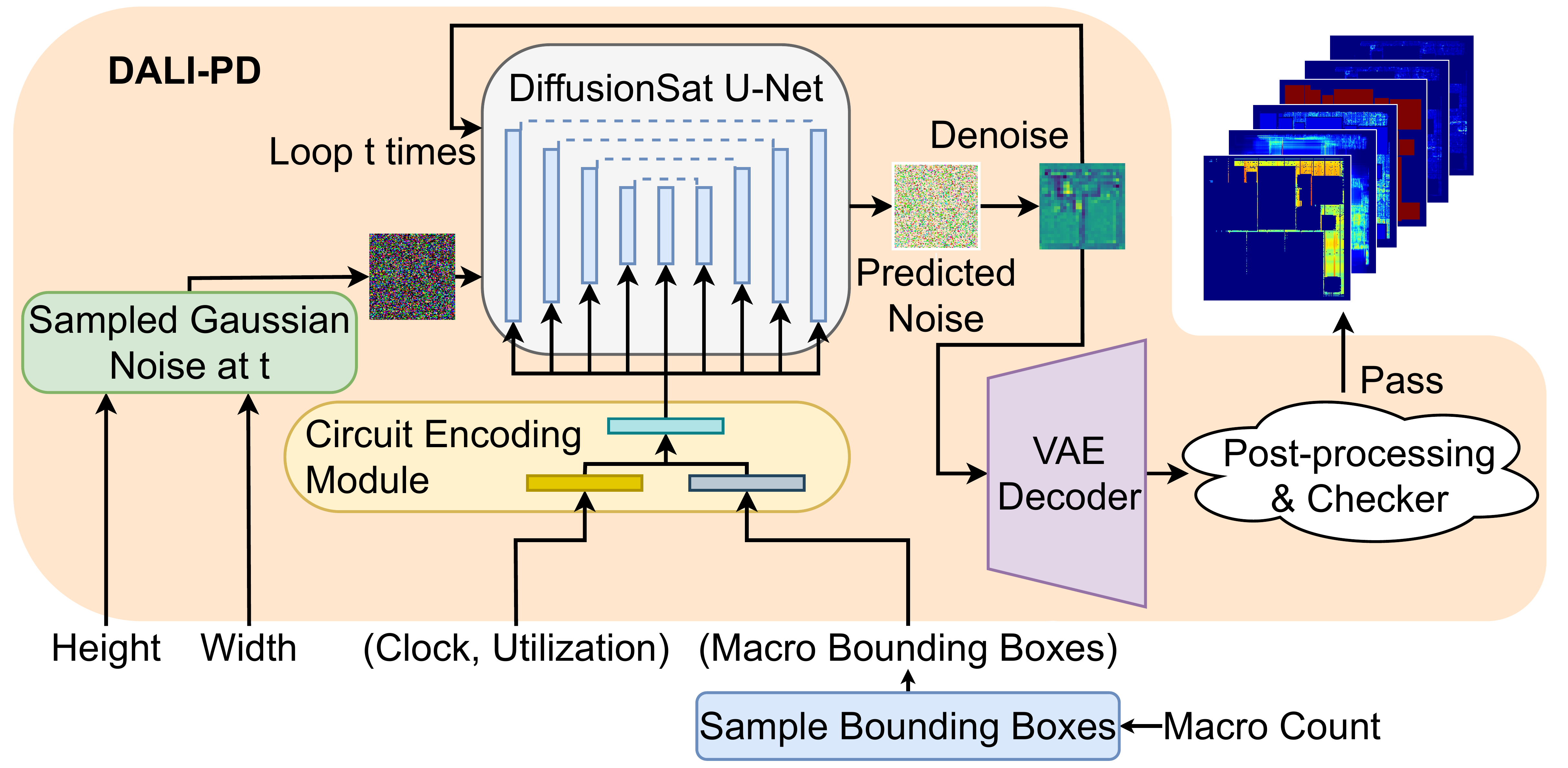}
    \vspace{-2.5mm}
    \caption{The proposed framework of DALI-PD.}
    \vspace{-4mm}
    \label{fig:inference}
\end{figure}

The DALI-PD framework for synthetic layout heatmaps generation is illustrated in Fig.~\ref{fig:inference}, where it randomly samples Gaussian noise and generates six categories of circuit layout heatmaps based on the provided circuit layout-related conditions, including the required clock period, layout height and width, utilization, and macro bounding boxes. DALI-PD has three modules: (1) the circuit encoding module, (2) the custom variational autoencoder (VAE), and (3) the diffusion model, which are each described below. Our goal is to learn the parametric probability distributions of the VAE and the latent diffusion U-Net, and use them to generate diverse heatmaps rapidly.

\subsection{Circuit encoding module}
\label{sec:embedding}
To guide the generation of circuit layout heatmaps, we provide a layout-related condition to the U-Net, as shown in Fig.~\ref{fig:inference} (yellow box). This condition encodes the CircuitNet heatmap features, for each data point $d_i \in D$, where $D$ is the CircuitNet dataset containing $N$ data points. We extract the clock period $c$, the utilization $u$, and the coordinates of the bounding boxes of macros $\forall b_j = (x_l^j{,}y_l^j{,}x_u^j{,}y_u^j) \in B$, where $B$ represents the macros in the design, $j = 1{,}\dots{,}M$, and $M$ is the number of macros in the design.

We transform the features into circuit embeddings $L = \{l_1{,}l_2{,}\dots{,}l_k\} \in \mathbb{R}^{k \times d_L}$, where $k \geq M$, using the learnable non-linear projection matrices $W_{cu} \in \mathbb{R}^{2 \times d_L}$ and $W_b \in \mathbb{R}^{4 \times d_L}$, and then combine them as shown in Eq.\eqref{eq:encode1}:

\vspace{-3mm}
\begin{equation}
\label{eq:encode1}
\small B_L = bW_b,\quad  \text{CU}_L = (c, u)W_{cu},\quad L = B_L + \text{CU}_L
\end{equation}
\vspace{-6mm}

To support a variable number of macros across data points, we pad $L $ to a fixed length $k$  by setting $b_{j+1:k} = (0, 0, 0, 0)$ and repeating $(c, u)$ $k$ times. Additionally, to handle varying heatmap dimensions, we normalize the macro bounding boxes by the height and width of the heatmap, respectively. With the conditioned circuit embedding, the latent diffusion model, described in Sec.~\ref{sec:unet}, can learn how to denoise the latent representation while preserving the characteristics of the circuit layout with the provided circuit condition.

\subsection{Variational autoencoder}
\label{sec:vae}

The combination of a VAE and a latent diffusion U-Net is widely adopted~\cite{2025iclrvaediffusion,2022diffusevae,si2023freeu,2024diffusionsat}. The VAE encoder downsamples the input to reduce the diffusion U-Net's computational cost by operating in a lower-dimensional latent space. The decoder reconstructs the image from the denoised latent produced by the U-Net. We adopt the same design in DALI-PD. While most methods use pretrained VAE decoders~\cite{rombach2022highresolutionimagesynthesislatent,esser2020taming} for RGB images, DALI-PD generates six types of circuit layout heatmaps, resulting in a six-channel output, twice that of standard RGB. Thus, RGB-based VAEs cannot be directly used.

\begin{figure}[t]
    \centering
    \includegraphics[width= 0.85\linewidth]{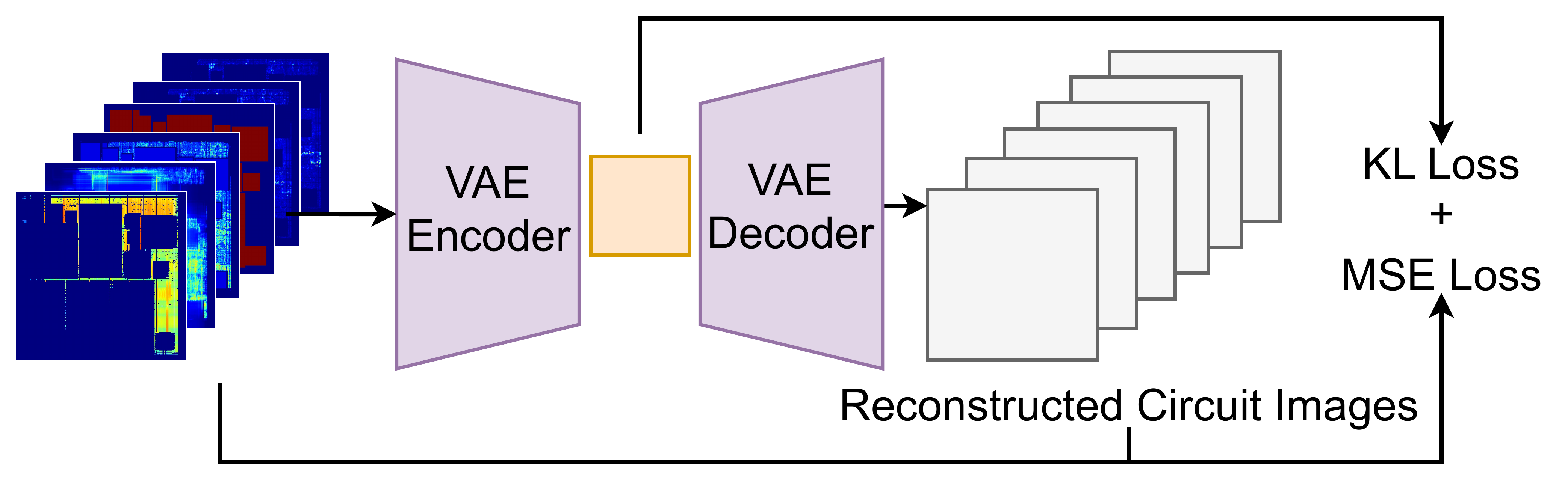}
    \vspace{-2.5mm}
    \caption{Training scheme of our variational autoencoder (VAE).}
    \vspace{-5mm}
    \label{fig:vae}
\end{figure}

To train the decoder to generate six categories of circuit layout heatmaps, the encoder must be jointly trained. As shown in Fig.\ref{fig:vae}, the VAE encoder compresses the six-channel layout heatmaps into a $d_{latent}$-channel latent representation, from which the decoder reconstructs the heatmaps. Both encoder and decoder consist of ResNet blocks with downsampling and upsampling components, respectively. To enable reconstruction, we maximize the log-likelihood of recovering input $x$ from latent $z$. Since integrating over the latent distribution $p(z)$ is intractable, we instead optimize the evidence lower bound (ELBO) objective\cite{2023nipselbo}.
ELBO indirectly optimizes the log-likelihood objective and includes a KL divergence term to regularize the encoder’s output toward a Gaussian distribution.

\subsection{Diffusion U-Net pretrained on satellite images}
The next step in DALI-PD is the diffusion model, which takes the circuit encoding and sampled Gaussian noise as input to generate a denoised latent image, which is then passed to the VAE decoder.

\subsubsection{Transfer-learning from satellite images}
\label{sec:transferlearning}
Similar to the approach in\cite{2021began}, which showed the effectiveness of leveraging the pre-trained weight of a GAN trained using urban satellite images, DALI-PD leverages the same key idea and utilizes DiffusionSat U-Net model~\cite{2024diffusionsat}, a latent diffusion model trained with satellite images specifically. We find that using a model with weights that have been pretrained with  satellite images generate more realistic layout heatmaps as we will show in Section~\ref{sec:results}.

\begin{figure}[t]
    \centering
    \vspace{-7mm}
    \includegraphics[width= 0.85\linewidth]{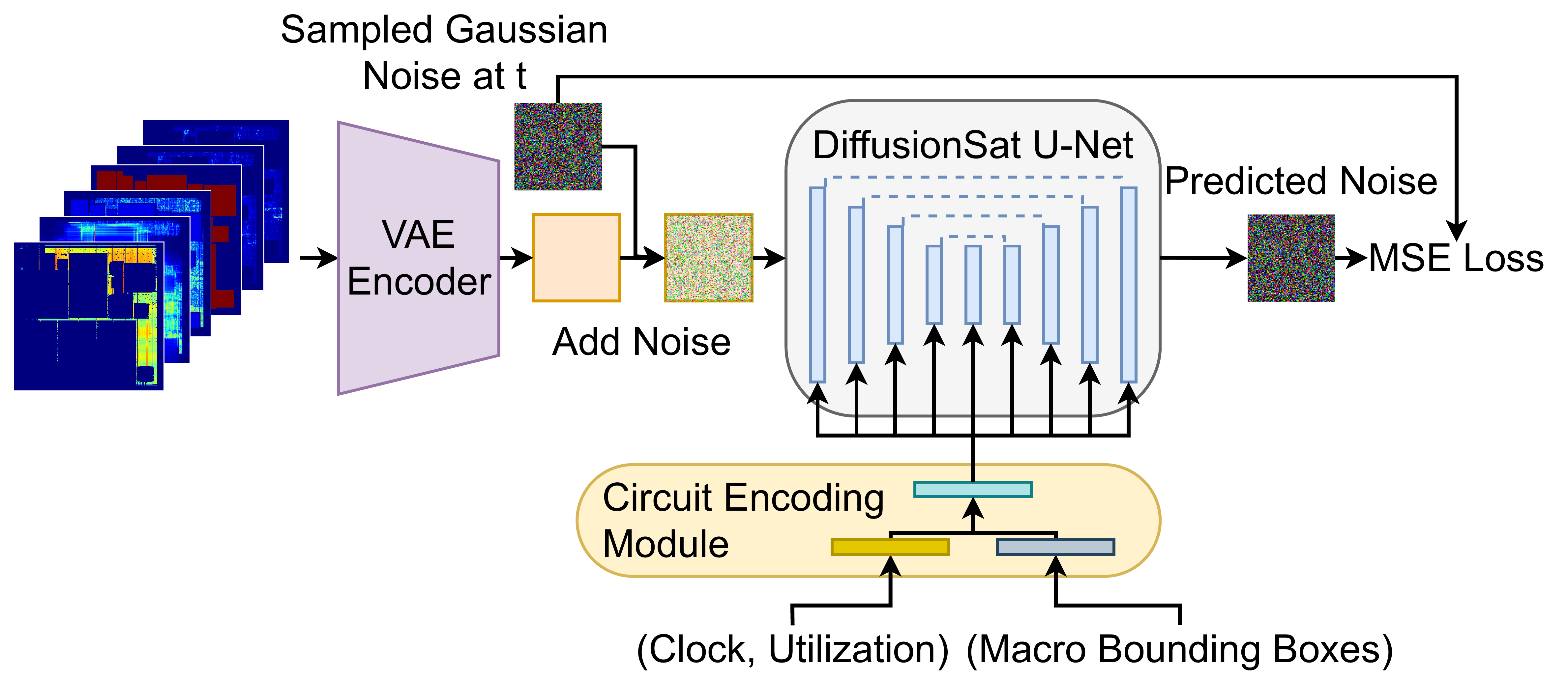}
    \vspace{-2.5mm}
    \caption{Training scheme of our latent diffusion Sat U-Net, which is pretrained with satellite images.}
    \vspace{-5.5mm}
    \label{fig:unet}
\end{figure}

\subsubsection{Conditioned denoising process}
\label{sec:cross_attn}
DiffusionSat is trained on text-image pairs and uses a CLIP-based text tokenizer~\cite{radford2021learningtransferablevisualmodels} to convert human prompts into text embeddings $\hat{L} \in \mathbb{R}^{\hat{k}\times \hat{d_L}}$ that share latent distributions with images. Here, $\hat{L}$ is the CLIP embedding, $\hat{k}$ the token sequence length, and $\hat{d_L}$ is the embedding dimension. These embeddings condition the latent images via a cross-attention mechanism~\cite{2017attention}, as shown below.

\vspace{-3.5mm}
\begin{equation}
\label{eq:cross_attn1}
\small Q = x_\text{latent}W_Q \in \mathbb{R}^{N \times d_\text{attn}}, W_Q \in \mathbb{R}^{F \times d_\text{attn}}
\end{equation}
\begin{equation}
\label{eq:cross_attn2}
\small K = \hat{L}W_K \in \mathbb{R}^{\hat{k} \times d_\text{attn}}, W_K \in \mathbb{R}^{ \hat{d_L} \times d_\text{attn}}
\end{equation}
\begin{equation}
\label{eq:cross_attn3}
\small V = \hat{L}W_V \in \mathbb{R}^{\hat{k} \times d_\text{attn}}, W_V \in \mathbb{R}^{ \hat{d_L} \times d_\text{attn}}
\end{equation}
\begin{equation}
\label{eq:cross_attn4}
\small \text{Attention}(Q, K, V) = \text{softmax}(\frac{QK^T}{\sqrt{d_\text{attn}}})V
\end{equation}
\vspace{-4mm}

\noindent
where $Q$, $K$, and $V$ represent the embeddings of query, key, and value in the cross-attention mechanism. $W_Q$. $W_K$, and $W_V$ are the attention matrices to convert the latent image $x_\text{latent}$ and the CLIP-based embedding $\hat{L}$ to $Q$, $K$, and $V$. $d_\text{attn}$ is the dimension of embeddings. $N$ is the product of the height and the width of the latent image $x_\text{latent}$, and $F$ is the dimension of the channel of the latent image $x_\text{latent}$. Since CircuitNet training dataset does not have text prompt for each heatmap, we use the circuit embedding introduced in Sec.~\ref{sec:embedding} to condition the circuit features into latent images in cross attention. To keep the original structure of the DiffusionSat U-Net, we intentionally set $d_L$ to match $\hat{d_L}$.

\subsubsection{Latent diffusion U-Net training}
\label{sec:unet}
The training process of the DiffusionSat-based latent diffusion U-Net is shown in Fig.~\ref{fig:unet}. After the VAE encoder generates a $d_\text{latent}$-channel downsampled image $x_0$, Gaussian noise $\epsilon_{\hat{\theta}}(x_t, t)$ is sampled with random time step $t$ and added to form $x_t$. The latent diffusion U-Net then uses both the skip-connection mechanism and cross-attention to incorporate external features, which in DALI-PD are circuit embeddings, to predict the noise $\epsilon_{\hat{\theta}}(x_t, t)$ added to $x_t$ at time step $t$. This predicted noise is removed from the sampled $x_t$ to denoise it and recover the original latent representation $x_0$. $d_\text{latent}$ is set to $4$ following the setting in~\cite{2024diffusionsat}.

Similar to training a VAE, the latent diffusion model is trained to learn the reverse process, i.e., reconstructing the clean latent image $x_0$ from a noisy version $x_t$ by optimizing ELBO~\cite{2023nipselbo}.




During inference, the model starts from a noise sample $x_T \sim \mathcal{N}(0, \text{I})$ and gradually denoises it using the reverse process. At each time step $t$, the model predicts the noise $\epsilon_{\hat{\theta}}(x_t, t)$, and the denoised sample $x_{t-1}$ is formed using the DDPM~\cite{ho2020denoising} reverse process:

\vspace{-4mm}
\begin{equation}
\label{eq:reverse_process}
\small x_{t-1} = \frac{1}{\sqrt{\alpha_t}} \left( x_t - \frac{1 - \alpha_t}{\sqrt{1 - \bar{\alpha}_t}} \epsilon_{\hat{\theta}}(x_t, t) \right) + \sigma_t \hat{z}, \quad \hat{z} \sim \mathcal{N}(0, \text{I})
\end{equation}

\noindent
where $\hat{z}$ is the standard Gaussian noise used to reintroduce randomness into the sampling process, and $\bar{\alpha}_t$ denotes the cumulative product of the noise schedule coefficients: $
\bar{\alpha}_t = \prod_{s=1}^{t} \alpha_s $. It represents the proportion of the original signal $x_0$ preserved at time step $t$ during the forward diffusion process. As $t$ increases, $\bar{\alpha}_t$ decreases, indicating that more noise has been added and less of the original signal remains.

\subsection{Heatmap generation via inference}
\label{sec:inference}
\underline{Inference using VAE and diffusion U-Net model} Once the models are trained, we perform inference as shown in Fig.~\ref{fig:inference}. Unlike training, where the layout height and width are fixed by the CircuitNet dataset, inference allows user-specified height and width, enabling variable aspect ratios and increasing layout diversity. To further enhance diversity, normalized macro bounding boxes $B$ are randomly sampled based on user-defined utilization $u$ and macro count $M$. These, along with the clock period $c$, are input to the circuit encoding module to generate circuit embeddings, which condition the denoising process.

The latent diffusion U-Net begins with Gaussian noise $x_T$ and iteratively denoises it using the DDPM reverse process. At each step $t$, the model predicts the added noise $\epsilon_{\hat{\theta}}(x_t, t)$ and computes $x_{t-1}$ according to Eq.\eqref{eq:reverse_process}. After $T$ steps, the final latent $x_0$ is obtained. This denoising is guided by cross-attention using circuit features (see Sec.\ref{sec:unet}). The VAE decodes $x_0$ into six layout heatmaps.

\noindent
\underline{Post-processing and sanity checker.} The VAE decoder output is passed through a post-processing module and a checker to ensure layout quality. Such approaches
are commonly adopted~\cite{rombach2022highresolutionimagesynthesislatent} to filter out low-quality results. Post-processing heuristics refine macro edges and align cells to power maps, correcting distortions from the diffusion model to produce more "circuit-like" heatmaps. Some outputs with overly sharp or overlapping macro boundaries are discarded. A checker then verifies whether the refined heatmaps meet design constraints—particularly utilization, which may drop during post-processing. For each parameter setting (height, width, clock period, utilization, macro count), DALI-PD iteratively samples until the checker approves the result. Despite this loop, generation remains fast, with inference taking 1–2 seconds per sample.

\section{Experimental Setup for DALI-PD evaluation}
\label{sec:results}

\subsection{DALI-PD training and test set description}
\label{sec:setup}

To evaluate DALI-PD’s ability to generate realistic circuit layout heatmaps for unseen circuit parameter sets—height, width, clock rate, utilization, and macro count, we setup the following experiment:
\begin{enumerate}
    \item We reserve four designs from the CircuitNet 28nm dataset exclusively for training, while the remaining two designs are used for testing, as shown in Table~\ref{tab:design_split}.
    \item We remove one hyperparameter option from each category of circuit parameters and include only those options in the test set.
\end{enumerate}

\renewcommand{\arraystretch}{1.5} 

\begin{table}[t]
\vspace{-6mm}
\caption{Training and test sets split of the six CircuitNet designs.}
\vspace{-1.75mm}
\centering
\fontsize{6.15pt}{7.5pt}\selectfont
\begin{tabular}{|c|c|c|c|c|}
\hline
 & \textbf{Design} & \textbf{Macro Count} & \textbf{Util (\%)} & \textbf{Clock (ns)} \\
\hline
\multirow{2}{*}{\shortstack{\textbf{Training} \\ \textbf{Set}}} 
 & RISCY-a, RISCY-FPU-a        & 4, 5    & \multirow{2}{*}{\shortstack{70, 75 \\ 80, 85}} & \multirow{2}{*}{2, 5} \\
\cline{2-3}  
 & RISCY-b, RISCY-FPU-b       & 13, 14   &                                          &                        \\
\hline
\multirow{2}{*}{\textbf{Test Set}}     
 & zero-riscy-a  & 3                         & \multirow{2}{*}{90}                      & \multirow{2}{*}{20}    \\
\cline{2-3}
 & zero-riscy-b  & 15                        &                                          &                        \\
\hline
\end{tabular}
\vspace{-7mm}
\label{tab:design_split}
\end{table}

Based on this, we build a training set of 2,861 unique data points from multiple synthesis and P\&R runs, and a test set of 67 points from the two unseen designs. We apply data augmentation (e.g., flipping, rotation) to expand the training set 11×, resulting in 34,332 samples. This is constructed using CircuitNet 28nm dataset.  All three modules in DALI-PD are trained solely on the training set and have no access to test-time circuit parameters. This setup allows us to assess DALI-PD's generalization and compare the generated heatmaps' realism against those in CircuitNet.

\subsection{DALI-PD training and hyperparameters}
\label{sec:model_training}
Using the training set, we train the VAE separately while the latent diffusion U-Net and the circuit encoding modules are jointly trained using 140K steps, with the U-Net weights initialized using DiffusionSat's pretrained weights, as described in Sec.~\ref{sec:transferlearning}. The VAE requires significantly more training steps 550K to converge, as it is trained from scratch. In contrast, the other two modules converge faster, demonstrating the effectiveness of transfer learning.
We train all three modules using a single NVIDIA A6000 GPU with the AdamW optimizer. To mitigate the effects of single-batch training, we adopt an exponential moving average (EMA)  model~\cite{moralesbrotons2024exponentialmovingaverageweights}.

\subsection{Evaluation techniques}
To evaluate the quality of our dataset in terms of both realism and diversity, we conduct two types of assessments. First, we use computer vision metrics to measure the statistical similarity between heatmaps generated by DALI-PD and those from the CircuitNet dataset, comparing both against a held-out test set. Second, we evaluate the utility of the dataset by training models on a downstream machine learning task: RUDY map prediction and IR drop prediction. Separate models are trained using the CircuitNet and DALI-PD datasets, and their performance is compared on a common test set to assess predictive accuracy.

\noindent
{\bf (1) Statistical metrics.} We use the following metrics to compare the generated images against the test set for realism and diversity. The mean and standard deviation of the generated heatmap distributions. We also compare the histograms of the 2D distributions of the test set and DALI-PD-generated image using the following: 

\noindent
\textit{ (A) The Fréchet inception distance (FID)} is a commonly used metric to evaluate the quality of generated images. FID measures the distance between the feature distributions of real images and generated images, using the activations of a specific layer (usually the pool3 layer) of the Inception v3 network. Lower FID values indicate that the generated data distribution is closer to the real data distribution. 



 

\noindent
\textit{(B) Structured similarity index measure (SSIM)}  is a perceptual metric that compares two images based on brightness, contrast, and structural patterns. It produces a score between 0 and 1, where 1 means the images look nearly identical. SSIM is used to detect duplicates or assess image quality in a way that aligns with visual perception.





\noindent
\textit{(C) L1 loss} is the pixel-wise mean absolute difference, which directly measures the distributional discrepancy. A low $L_1$ score indicates that the generated sample has a similar spatial distribution. 

\noindent
{\bf (2) Downstream ML applications for evaluation.} To comprehensively test the diversity and realism of DALI‑PD–generated we test whether knowledge learned from the DALI‑PD-generated heatmaps transfers to real circuits in the CircuitNet dataset~\cite{2023circuitnet} by performing two ML tasks: (A) IR drop prediction using cell density, power, and toggle‑rate–scaled‑power heatmaps as features; and (B) RUDY prediction using cell density and macro region heatmaps as features. For both these tasks we use a U-Net similar to existing work on IR drop~\cite{iredge, maveric} and congestion prediction~\cite{kim2021machine}. The model architecture and hyperparameters for the two tasks are  in Table~\ref{tab:unet_arch}.

\begin{table}[t]
\centering
\vspace{-4.2mm}
\caption{Evaluation model architecture and training configuration.}
\vspace{-1.5mm}
\renewcommand{\arraystretch}{1} 
\fontsize{6.15pt}{7.5pt}\selectfont
\begin{tabular}{|>{\centering\arraybackslash}m{1.3cm}
                |>{\centering\arraybackslash}m{1.15cm}
                |>{\centering\arraybackslash}m{1.15cm}
                |>{\centering\arraybackslash}m{0.9cm}
                |>{\centering\arraybackslash}m{1.0cm}
                |>{\centering\arraybackslash}m{0.6cm}|}
\hline
\shortstack[c]{\textbf{Architecture}} &
\shortstack[c]{\textbf{Conv}\\\textbf{Kernel Size}} &
\shortstack[c]{\textbf{Deconv}\\\textbf{Kernel Size}} &
\shortstack[c]{\textbf{Channels}} &
\shortstack[c]{\textbf{Learning}\\\textbf{Rate}} &
\shortstack[c]{\textbf{Batch}\\\textbf{Size}} \\ \hline
\shortstack[c]{U-Net} &
\shortstack[c]{5} &
\shortstack[c]{5} &
\shortstack[c]{16, 64,\\128, 512} &
\shortstack[c]{$5 \times 10^{-5}$} &
\shortstack[c]{64} \\ \hline
\end{tabular}
\vspace{-3.5mm}
\label{tab:unet_arch}
\end{table}

\noindent
{\bf (3) Baseline to justify DiffusionSat model.} 
To evaluate the impact of using DiffusionSat, as an ablation study, we introduce a baseline latent diffusion U-Net trained with the same configuration. The key difference is in the initialization: the baseline is fine-tuned directly from Stable Diffusion, while the DALI-PD model is fine-tuned from DiffusionSat~\cite{2024diffusionsat}, which builds on Stable Diffusion with pretraining.



\begin{figure}[t]
    \centering
    \vspace{-0mm}
    \includegraphics[width= 0.95\linewidth]{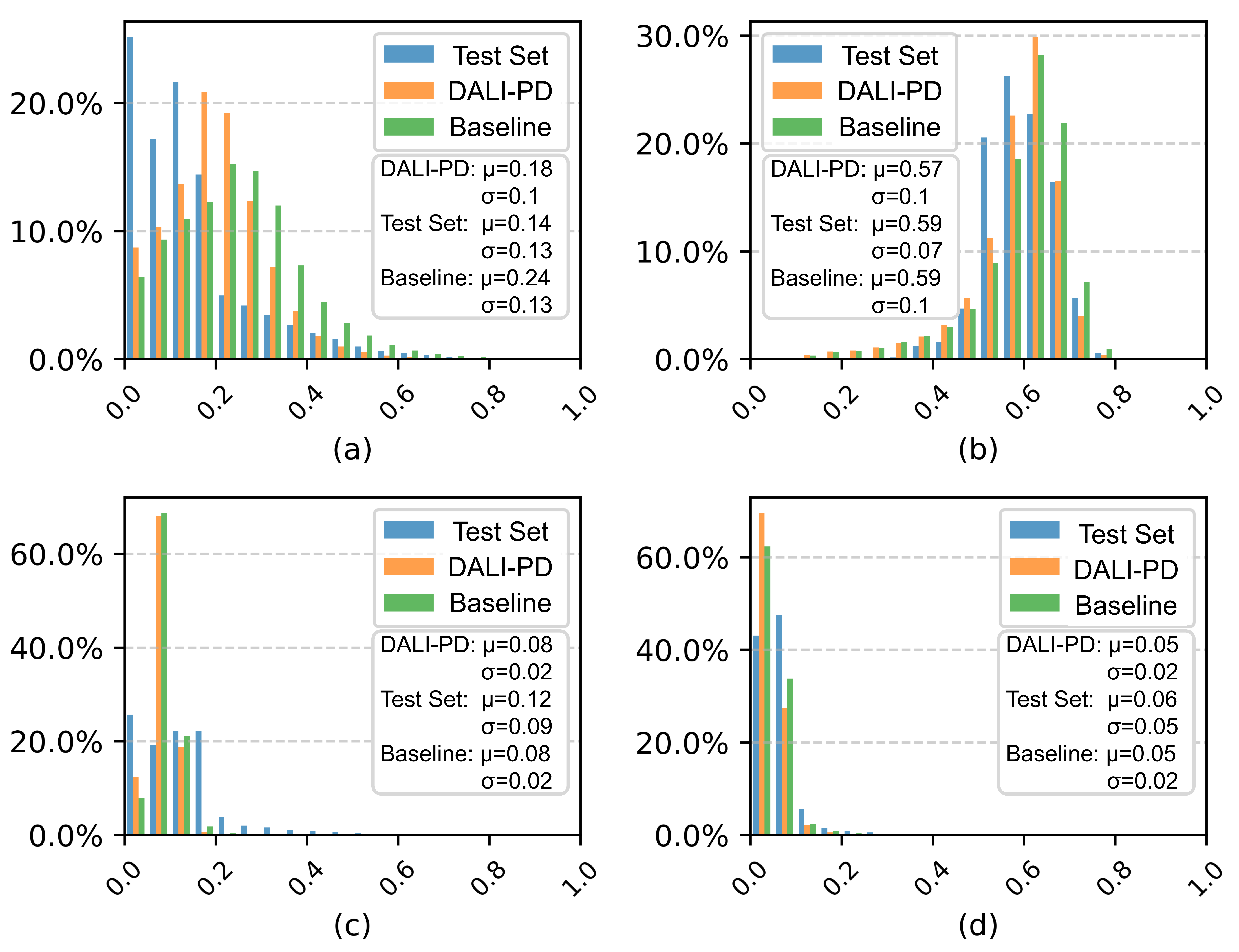}
    \vspace{-3mm}
    \caption{Comparison of (a) RUDY, (b) IR drop, (c) power, and (d) toggle rate-scaled power distributions across CircuitNet test set, DALI-PD, and baseline.}
    \vspace{-3.9mm}
    \label{fig:baseline_our_test}
\end{figure}


\section{Evaluation of DALI-PD}
We extract circuit parameters from the test set, including the height and width of the layout, clock period, utilization, and bounding boxes of the macros, to generate layout heatmaps. For sampling, we use the DDPM sampler with $100$ steps and a guidance scale of $1.0$. We generate over 20,000 data points, which form the DALI-PD dataset. 

\subsection{Statistical comparison of DALI-PD-generated heatmaps}
\label{sec:test_set}
Using statistical metrics, we evaluate the dataset for both realism (similarity to the test set) and diversity. 
For diversity, we ensure that the generated dataset does not contain two maps that are similar to each other by examining pairwise comparisons of SSIM. We compare the SSIM values within our dataset and within the CircuitNet dataset to verify that our dataset is sufficiently distinct.

\begin{figure*}[t]
    \centering
    \vspace{-6mm}
    \includegraphics[width= \linewidth]{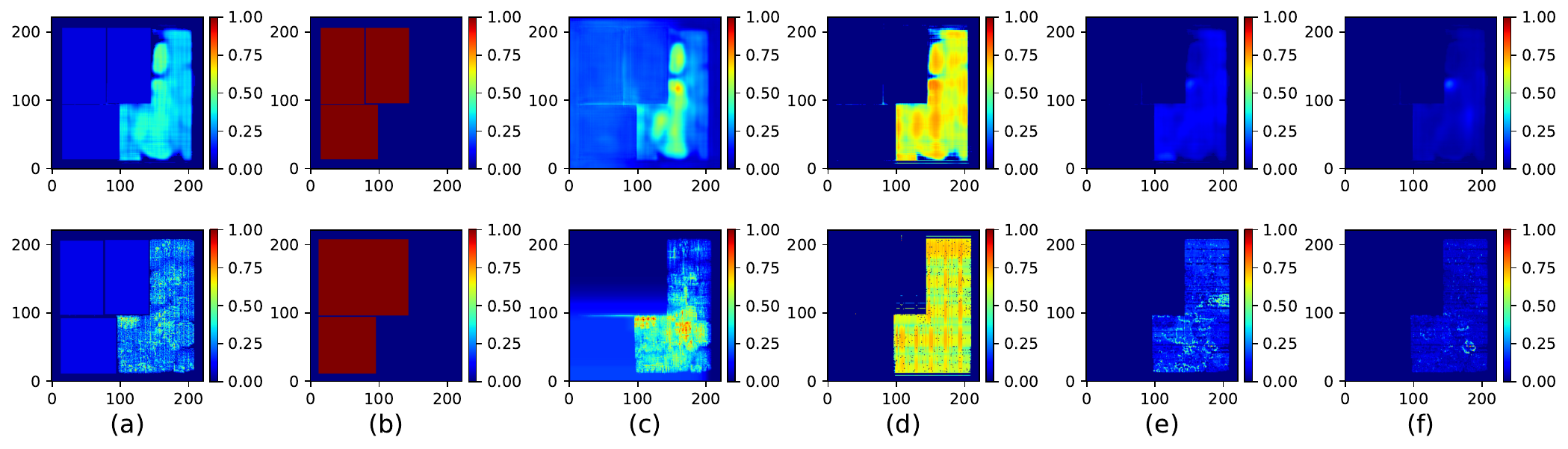}
    \vspace{-5.75mm}
    \caption{The first row shows example DALI-PD-generated heatmaps, conditioned on circuit features from the second row—real heatmaps from the ``zero-riscy-a" design in the CircuitNet test set. The six heatmaps are: (a) cell density, (b) macro region, (c) RUDY, (d) IR drop, (e) power, and (f) toggle-scaled power.}
    \vspace{-2mm}
    \label{fig:example}
\end{figure*}

\noindent
\underline{Realism} For realism, we compare the generated dataset and the CircuitNet real circuit test set. Specifically, we examine the mean, standard deviation, percentage hotspots,  FID, and make visual comparisons between the test set and generated heatmaps.  

\noindent
(1) Histogram:  The histogram of the heatmap distributions of the $67$ data points in the test set is compared against the DALI-PD generated circuit layout heatmaps as shown in Fig.~\ref{fig:baseline_our_test}. Our generated heatmaps have histogram distributions similar to those of the test set from CircuitNet for IR drop (Fig.~\ref{fig:baseline_our_test}(b)) and power (Fig.~\ref{fig:baseline_our_test}(c) and (d)). The histogram distribution of the RUDY heatmaps in Fig.~\ref{fig:baseline_our_test}(a) is slightly skewed, as this property is easily affected by the geometry of the layout itself. However, the overall mean and standard deviation, as shown in Fig.~\ref{fig:baseline_our_test}, remain within a reasonable margin.  We evaluate the same mean and standard deviation metrics using the baseline model. These results show that our generative framework, whether utilizing the baseline U-Net or the DiffusionSat based version, can produce similar mean and standard deviation of the histograms when compared to the real CircuitNet test set. However, mean and standard deviation being similar is not enough; we also compare the FID of these histograms for the six categories of heatmaps. 

\noindent
(2) FID comparison against baseline: To evaluate the effectiveness of the DiffusionSat model and the realism of our dataset, we use the FID score.  Table~\ref{tab:fid_iter_compact} presents the average FID score of the four (RUDY, IR drop, power, and toggle-scaled power) categories of the generated heatmaps using two different latent diffusion U-Nets, compared to the heatmaps in the test set, along with the number of iterations required to generate a set of heatmaps that pass the post-processing checker’s threshold. We observe that the baseline U-Net takes more than $8\times$ the number of iterations to generate valid heatmaps compared to DALI-PD, demonstrating that transfer learning from satellite-image-based knowledge enhances stability. From Table~\ref{tab:fid_iter_compact}, we can observe that the baseline has a higher average FID score, which aligns with the observed instability in generation quality, proving the effectiveness of transfer learning from satellite image knowledge.
\begin{table}[t]
\centering
\vspace{-4mm}
\caption{FID and iteration comparison between DALI-PD and baseline.}
\vspace{-2mm}
\fontsize{6.15pt}{7.5pt}\selectfont
\begin{tabular}{|l|crrr|crrr|}
\hline
\multirow{2}{*}{} & \multicolumn{4}{c|}{\textbf{FID}} & \multicolumn{4}{c|}{\textbf{Iteration}} \\ \cline{2-9} 
 &
  \multicolumn{1}{c|}{Avg} &
  \multicolumn{1}{c|}{Stdv} &
  \multicolumn{1}{c|}{Min} &
  \multicolumn{1}{c|}{Max} &
  \multicolumn{1}{c|}{Avg} &
  \multicolumn{1}{c|}{Stdv} &
  \multicolumn{1}{c|}{Min} &
  \multicolumn{1}{c|}{Max} \\ \hline
\multicolumn{1}{|c|}{\textbf{DALI-PD}} &
  \multicolumn{1}{r|}{41.86} &
  \multicolumn{1}{r|}{30.23} &
  \multicolumn{1}{r|}{0.07} &
  171.7 &
  \multicolumn{1}{r|}{2.68} &
  \multicolumn{1}{r|}{2.67} &
  \multicolumn{1}{r|}{1} &
  12 \\ \hline
\multicolumn{1}{|c|}{\textbf{Baseline}} &
  \multicolumn{1}{r|}{70.29} &
  \multicolumn{1}{r|}{39.71} &
  \multicolumn{1}{r|}{0.06} &
  199.31 &
  \multicolumn{1}{r|}{24} &
  \multicolumn{1}{r|}{18.1} &
  \multicolumn{1}{r|}{7} &
  117 \\ \hline
\end{tabular}
\vspace{-2.5mm}
\label{tab:fid_iter_compact}
\end{table}

\noindent
(3) Visual comparison: We plot a selected sample of heatmaps in Fig.~\ref{fig:example} generated by DALI-PD that has the highest SSIM to the test set of CircuitNet. The results shown in the first row of the figure are generated by DALI-PD. The second row is the test set from CircuitNet. It's important to note that this test set from CircuitNet was not within the training set of DALI-PD.  These distributions serve as a visual comparison of the realism of the generated heatmaps. The generated results largely follow the given circuit parameters. Since the generation process is not a deterministic parameter-to-image mapping, during each step of DDPM diffusing, different styles of heatmaps can be introduced while still preserving realistic circuit properties. Here we show the visual similarity between this sample and the CircuitNet ``zero-riscy-a" test data point.

\begin{table}[t]
\centering
\vspace{-2mm}
\caption{The first row shows \%low-density areas ($<0.1$). Other rows show \% hotspots ($>0.9$) for RUDY, IR drop, and power.}
\vspace{-2mm}
\fontsize{6.15pt}{7.5pt}\selectfont
\begin{tabular}{|c|r|r|}
\hline
\multicolumn{1}{|l|}{} & \multicolumn{1}{c|}{\textbf{Test Set}} & \multicolumn{1}{c|}{\textbf{DALI-PD}} \\ \hline
\textbf{Low Cell Density Area (\%)}   &  58.83  & 62.78  \\
\hline
\textbf{Cell Density Hotspot (\textperthousand{})}     &  0.15 &  0       \\
\hline
\textbf{RUDY Hotspot (\%)}                             & 1.2  &  0.68  \\
\hline
\textbf{IR Drop Hotspot (\%)}                          & 0.03  &  0.01  \\
\hline
\textbf{Power Hotspot (\textperthousand{})}            & 0.04  &  0      \\
\hline
\textbf{Scaled Power Hotspot (\textperthousand{})}     & 0.04  &  0      \\
\hline

\end{tabular}
\vspace{-6mm}
\label{tab:hotspot}
\end{table}

\noindent
(4) Hotspot comparison:
Table~\ref{tab:hotspot} shows the percentage of hotspot regions (values $>0.9$) and non-congested areas (values $<0.1$) in the heatmaps. Only the cell density maps have non-congested regions, while all others do not, in both the test and generated sets. This indicates that the generated heatmaps closely match the hotspot and cell density distributions of real data. As DALI-PD operates on normalized heatmaps, the outputs can be rescaled to user-defined ranges—for example, IR drop maps can represent any voltage range.

\noindent
(5) L1 loss:  Table~\ref{tab:swd_l1} reports the average L1 score between DALI-PD-generated heatmaps and real CircuitNet heatmaps in the test set. The L1 score is computed as the pixel-wise absolute difference between each generated image and its corresponding real image. Specifically, we use 67 test set images from CircuitNet, generate 67 corresponding heatmaps using DALI-PD with matching input circuit parameters, and compute the L1 loss across these pairs. A lower average L1 score indicates that the generated images are structurally closer to the real dataset. While the L1 score is not zero—reflecting that the images are not identical—the values remain within a reasonable range, averaging under 26\% (0.26 for average macros). Given that the mean and standard deviation of the image histograms are closely aligned with those of the real test set (as shown in Fig.~\ref{fig:baseline_our_test}), and the FID scores are lower than the baseline, these L1 differences suggest that DALI-PD generates diverse yet realistic heatmaps.



\begin{table}[t]
\vspace{-4mm}
\caption{L1 loss between DALI-PD and CircuitNet test heatmaps. Lower means lesser pixel-wise differences (more realism).}
\vspace{-2mm}
\centering
\fontsize{6.15pt}{7.5pt}\selectfont
\begin{tabular}{|c|c|c|c|c|c|c|}
\hline
 & \textbf{Cell} & \textbf{Macro} & \textbf{RUDY} & \textbf{IR drop} & \textbf{Power} & \textbf{Scaled power} \\
\hline
\textbf{avg} & 0.09 & 0.26 & 0.17 & 0.17 & 0.04 & 0.02 \\
\hline
\textbf{stdv} & 0.02 & 0.17 & 0.04 & 0.08 & 0.02 & 0.01 \\
\hline
\end{tabular}
\vspace{-3mm}
\label{tab:swd_l1}
\end{table}

\noindent
\underline{Large diverse dataset and runtimes} We use t-SNE for visualization and SSIM to assess diversity, ensuring all data points are distinct.

\begin{figure}[t]
    \centering
    \vspace{-5mm}
    \includegraphics[width= 0.58\linewidth]{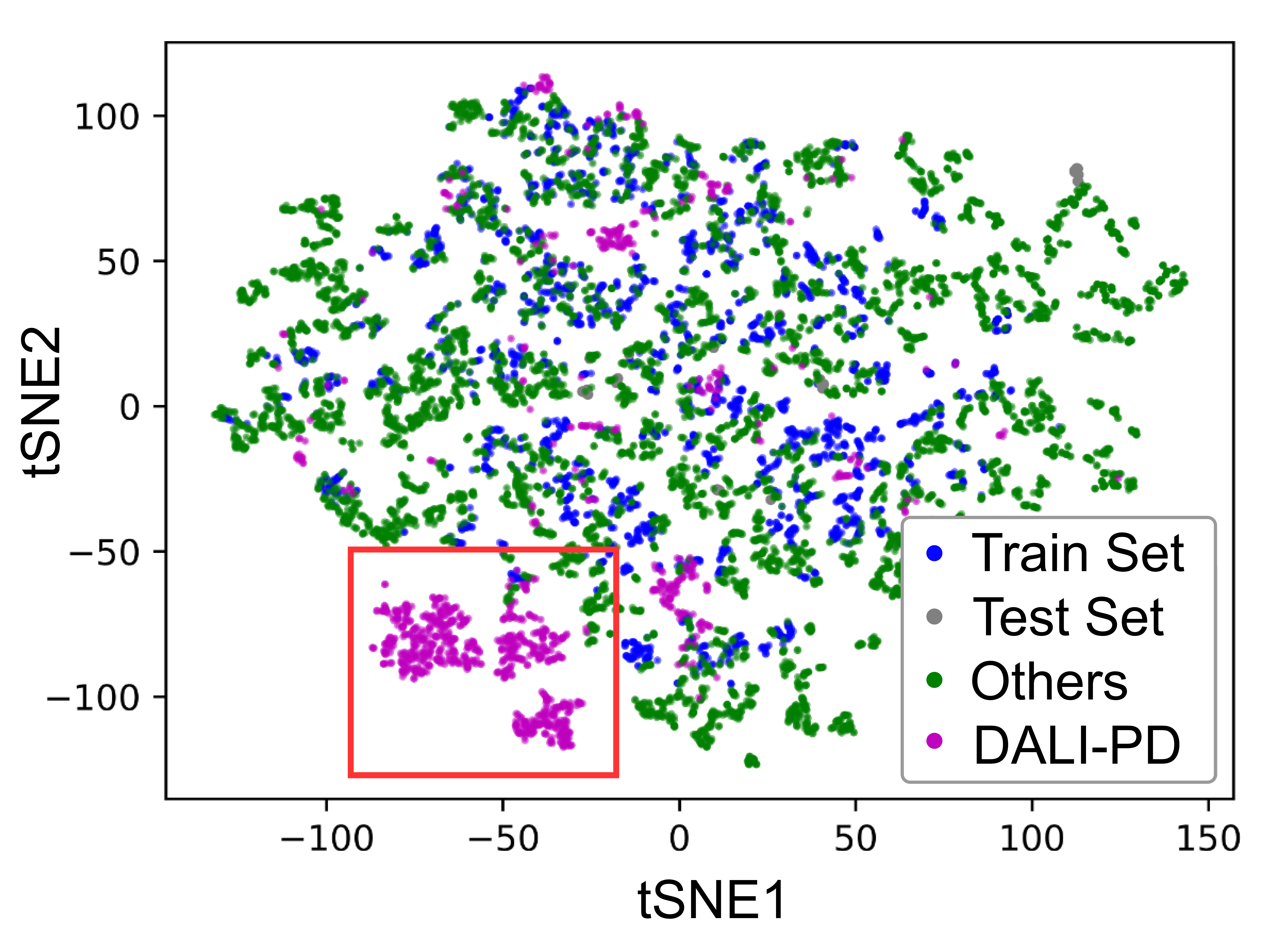}
    \vspace{-3mm}
    \caption{The t-SNE plot shows the correlation between different heatmaps, with the red box highlighting the diversity introduced by DALI-PD.}
    \label{fig:tsne}
    \vspace{-2mm}
\end{figure}

\noindent
(1) t-SNE score: Using DALI-PD, we open-source a large dataset of 23,070 layout heatmaps for each category~\cite{DALI-PD-GH-commit}. Fig.~\ref{fig:tsne} illustrates the data diversity introduced by DALI-PD. ``Others'' denotes data points not included in the train or test sets. We randomly sample 1,000 layout heatmaps across six categories from the dataset. For each, we extract the average, the standard deviation, and the hotspot percentage of RUDY, IR drop, power, scaled power, and aspect ratio to visualize the diversity using t-SNE. The figure highlights the diversity of the dataset as there is a new region in the 2D space in the red box that was previously not covered, which DALI-PD now covers.

\begin{figure}[t]
    \centering
    \vspace{-3mm}
    \includegraphics[width= 0.7\linewidth]{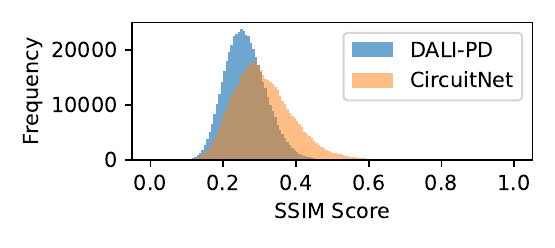}
    \vspace{-5mm}
    \caption{Average pairwise SSIM histograms across six categories for CircuitNet and DALI-PD, each computed using 1,000 sampled heatmaps.}
    \vspace{-5mm}
    \label{fig:ssim}
\end{figure}

\begin{table}[t]
\centering
\vspace{-1.45mm}
\caption{Pairwise SSIM score for all categories for CircuitNet and DALI-PD, each using 1,000 sampled heatmaps.}
\vspace{-2mm}
\fontsize{6.15pt}{7.5pt}\selectfont
\begin{tabular}{|l l|r|r|r|r|r|r|}
\hline
\multicolumn{2}{|c|}{\multirow{2}{*}{\textbf{SSIM}}} & 
\multirow{2}{*}{\shortstack{\textbf{Cell} \\ \textbf{Density}}} & 
\multirow{2}{*}{\shortstack{\textbf{Macro} \\ \textbf{Region}}} & 
\multirow{2}{*}{\textbf{RUDY}} & 
\multirow{2}{*}{\shortstack{\textbf{IR} \\ \textbf{Drop}}} & 
\multirow{2}{*}{\textbf{Power}} & 
\multirow{2}{*}{\shortstack{\textbf{Scaled} \\ \textbf{Power}}} \\

 &  &  &  &  &  &  &  \\
\hline
\multirow{2}{*}{\textbf{DALI-PD}} 
& \multicolumn{1}{|l|}{\textbf{Avg}}  & 0.233 & 0.246 & 0.233 & 0.250 & 0.288 & 0.318 \\
\cline{2-8}
& \multicolumn{1}{|l|}{\textbf{Stdv}} & 0.052 & 0.081 & 0.062 & 0.071 & 0.073 & 0.079 \\
\hline
\multirow{2}{*}{\textbf{CircuitNet}} 
& \multicolumn{1}{|l|}{\textbf{Avg}}  & 0.351 & 0.423 & 0.352 & 0.274 & 0.209 & 0.261 \\
\cline{2-8}
& \multicolumn{1}{|l|}{\textbf{Stdv}} & 0.089 & 0.088 & 0.137 & 0.072 & 0.094 & 0.096 \\
\hline
\end{tabular}
\vspace{-5mm}
\label{tab:ssim}
\end{table}

\noindent
(2) SSIM comparison: To ensure no two maps in the DALI-PD dataset are identical or similar, we conducted a pairwise SSIM evaluation on the same sample of 1000 data points in DALI-PD for each of the six categories of layout heatmaps. The number of unique pairwise comparisons among 1000 data points, excluding self-pairs and duplicate pairs, is given by the binomial coefficient: $\binom{1000}{2} = \frac{1000 \times 999}{2} = 499{,}500$. We plot the average SSIM across all six categorical heatmaps in a histogram in Fig.\ref{fig:ssim}. The histogram shows that DALI-PD has a low SSIM score with all pairwise metrics below 0.45. In contrast, the CircuitNet dataset has scores that go up to the value 0.6, indicating less diversification in CircuitNet compared to DALI-PD. Table~\ref{tab:ssim} shows the SSIM score for six heatmap categories for both CircuitNet and DALI-PD datasets.

\noindent
(2) Runtime: The full DALI-PD dataset of 23,070 samples was generated in just 4 hours using 8 NVIDIA L40S and 4 NVIDIA A6000 GPUs with 1-2 seconds of runtime per inference.  It is also over 2× larger than CircuitNet, which in contrast takes weeks to generate~\cite{2024circuitnet}.

\subsection{Effectiveness of the DALI‑PD dataset for downstream ML tasks}
To evaluate the effectiveness of the DALI-PD dataset for downstream ML tasks, we train U-Net models for IR drop and RUDY prediction. The configuration of the U-Net model is shown in Table~\ref{tab:unet_arch}. We train this model with $125$ steps using the DALI-PD dataset, which includes  $23,070$ layout configurations, and compare it against a model trained on $2,861$ heatmaps from CircuitNet~\cite{2023circuitnet}.

\begin{figure}[t]
    \centering
    \vspace{-2mm}
    \includegraphics[width= 0.84\linewidth]{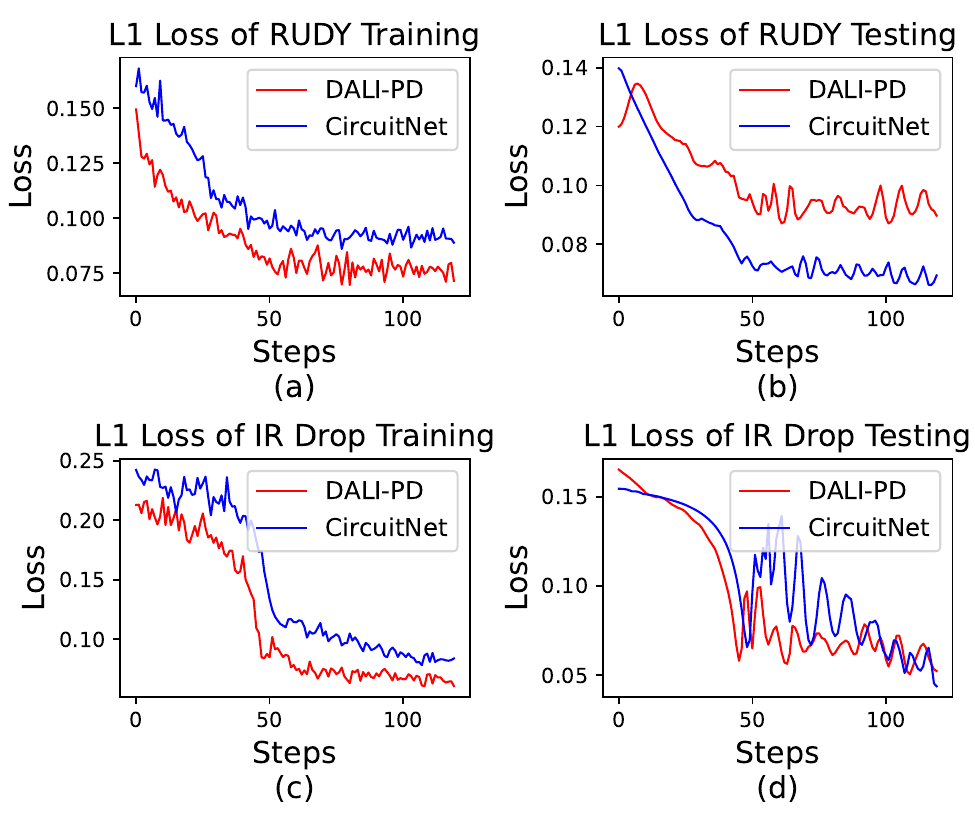}
    \vspace{-4.5mm}
    \caption{Training and testing loss of the U-Net model trained on CircuitNet and DALI-PD for RUDY and IR drop prediction. All models are evaluated on the same separate CircuitNet test set.}
    \vspace{-2mm}
    \label{fig:loss_without_mix}
\end{figure}

\begin{table}[t]
\centering
\fontsize{6.15pt}{7.5pt}\selectfont
\vspace{-2mm}
\caption{L1 loss of U-Net model trained on DALI-PD and CircuitNet datasets for IR drop and RUDY prediction tasks.}
\vspace{-2mm}
\begin{tabular}{|c|r|r|r|r|}
\hline
\multirow{2}{*}{} & \multicolumn{2}{c|}{\textbf{IR Drop}} & \multicolumn{2}{c|}{\textbf{RUDY}} \\ \cline{2-5} 
 & \multicolumn{1}{c|}{$L_1$} & \multicolumn{1}{c|}{Hotspot $L_1$} & \multicolumn{1}{c|}{$L_1$} & \multicolumn{1}{c|}{Hotspot $L_1$} \\ \hline
\textbf{CircuitNet}         & 0.036 & 0.085 & 0.067 & 0.202 \\ \hline
\textbf{DALI-PD}            & 0.062 & 0.14 & 0.092 & 0.168 \\ \hline
\end{tabular}
\vspace{-2mm}
\label{tab:accuracy}
\end{table}

\begin{figure}[t!]
    \centering
    \vspace{-1.5mm}
    \includegraphics[width= 0.89\linewidth]{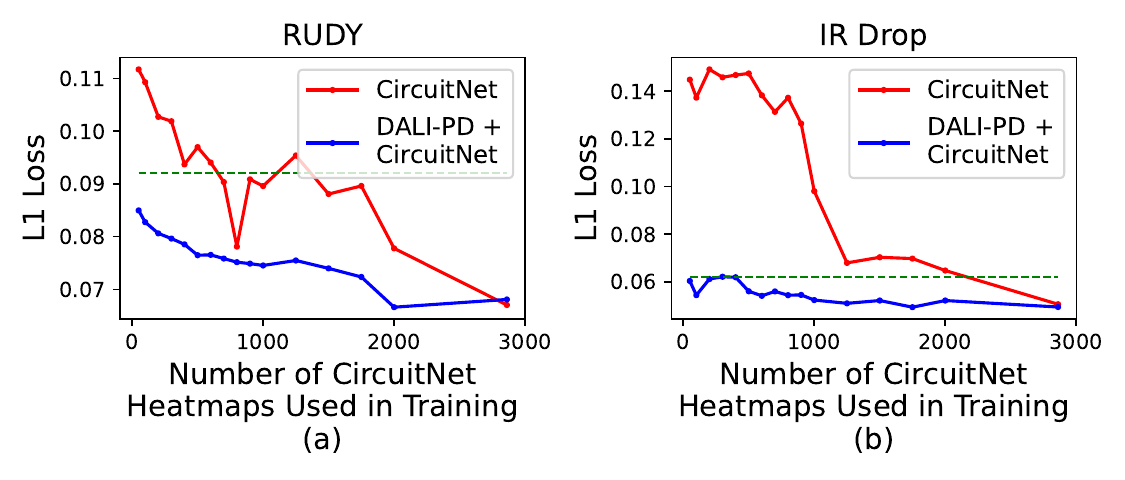}
    \vspace{-3.5mm}
    \caption{L1 loss for (a) RUDY and (b) IR drop prediction. The x-axis shows the number of real circuit heatmaps used for fine-tuning. The green line shows the L1 loss of the DALI-PD-pretrained model (see Table~\ref{tab:accuracy}).}
    \label{fig:data_amount_curve}
    \vspace{-4.5mm}
\end{figure}

We evaluate performance on the CircuitNet test set, with the train/test split described in Sec.~\ref{sec:setup}. The DALI-PD dataset is sampled from~\cite{DALI-PD-GH-commit}. Fig.~\ref{fig:loss_without_mix}(a) and (c) show the L1 loss convergence during training when the same model architecture and hyperparameters are used for training on CircuitNet and DALI‑PD, respectively for RUDY and IR drop prediction. Fig.~\ref{fig:loss_without_mix}(b) and (d) plot the L1 loss on the same test set throughout training.
From Fig.~\ref{fig:loss_without_mix}, we observe that both models converge well, indicating that meaningful representations are learned. The decreasing test losses demonstrate that training on either dataset is effective.  Table~\ref{tab:accuracy} summarizes the final L1 loss values. Training entirely on CircuitNet yields the best overall performance, while training solely on DALI‑PD achieves comparable but slightly accuracy. We also evaluate hotspot regions (top 10\% highest values): the CircuitNet-trained model performs better for IR drop prediction, while the DALI‑PD-trained model slightly outperforms it for RUDY. From Table~\ref{tab:accuracy}, we observe that models trained on real-world heatmaps generally perform better. However, preparing real-world datasets is costly, as each heatmap takes hours to generate~\cite{2024circuitnet}. 

To further evaluate the effectiveness of synthetic data, we make a case that we do not need as much real-circuit data when synthetic data is available to achieve the same accuracy.  We therefore use the model trained solely on synthetic DALI-PD data, and then fine-tune it with varying amounts of real CircuitNet data. The results are shown in Fig.~\ref{fig:data_amount_curve}. We find that with fewer than $500$ real-circuit heatmaps, models pre-trained on synthetic data outperform those trained on real-circuit data alone due to the diversity of the synthetic data.  Fine-tuning the synthetic-data-trained model with real heatmaps significantly boosts performance.  We conclude that when real-circuit heatmaps are limited, incorporating synthetic data is useful.

\section{Conclusion}
\label{sec:conclusion}

We propose DALI-PD, a fast generative framework for circuit heatmap generation, producing high-quality samples in seconds—far faster than traditional EDA flows. To demonstrate the effectiveness of our dataset, we show that DALI-PD generates diverse yet realistic layouts statistically and improves performance on downstream ML tasks such as RUDY and IR drop prediction. 

\clearpage
\bibliographystyle{misc/ieeetr2}
\bibliography{misc/bibfile}

\end{document}